\pdfoutput=1

\documentclass[11pt]{article}

\usepackage[]{ACL2023}

\usepackage{times}
\usepackage{latexsym}

\usepackage[T1]{fontenc}

\usepackage[utf8]{inputenc}

\usepackage{microtype}

\usepackage{inconsolata}

\usepackage{graphicx}
\usepackage{comment}
\usepackage{multirow}
\usepackage{booktabs}
\usepackage{subfigure}
\usepackage{lscape}
\usepackage{color,soul}
\usepackage{mathtools}
\usepackage{enumitem}
\usepackage{afterpage}

\title{Autoregressive Language Models for Knowledge Base Population: A case study in the space mission domain}

\author{Andres Garcia-Silva \\
  Language Technology Research Lab \\
  Expert.ai / Madrid, Spain\\
  \texttt{agarcia@expert.ai} \\\And
  Jose Manuel Gomez-Perez \\
  Language Technology Research Lab \\
  Expert.ai / Madrid, Spain \\
  \texttt{jmgomez@expert.ai} \\}

\begin{document}
\maketitle
\begin{abstract}
Knowledge base population KBP plays a crucial role in populating and maintaining knowledge bases up-to-date in organizations by leveraging domain corpora.  Motivated by the increasingly large context windows supported by large language models, we propose to fine-tune an autoregressive language model for end-to-end KPB. Our case study involves the population of a space mission knowledge graph. To fine-tune the model we generate a dataset for end-to-end KBP tapping into existing domain resources. Our case study shows that fine-tuned language models of limited size can achieve competitive and even higher accuracy than larger models in the KBP task. Smaller models specialized for KBP offer affordable deployment and lower-cost inference. Moreover, KBP specialist models do not require the ontology to be included in the prompt, allowing for more space in the context for additional input text or output serialization. 
\end{abstract}

\section{Introduction}
Knowledge graphs are data models to represent domain and world knowledge where entities and relationships are arranged in a graph structure \cite{vrandevcic2014wikidata, speer2017conceptnet}. Knowledge graphs have been used successfully in a variety of applications, including search engines, recommender systems \cite{guo2020recommendersurvey}, and intelligent assistants \cite{JiKGSurvey2022}. However, their success depends on an accurate and comprehensive population \cite{west_knowledge_2014,Galagarra2017}, which is often a labor-intensive process. 

Knowledge base population KBP \cite{ji_knowledge_2011} aims at automatically populating knowledge bases from text corpora. A typical KBP process is composed of several components \cite{mesquita2019knowledgenet} 
, such as entity linking \cite{wu_enriching_2019}, 
slot-filling  \cite{surdeanu2014slotfillingtac14} and relation extraction  \cite{adel2019type} 
. In this scenario errors can propagate through components in the overall workflow \cite{trisedya-etal-2019-neural}. Modern end-to-end KBP approaches have been proposed to avoid error propagation following a sequence to sequence approach \cite{sutskever2014seq2seq} where an autoregressive decoder 
\cite{trisedya-etal-2019-neural} or a non-autoregressive decoder \cite{sui-etal-2021-set} processes the input sentence and produces the triples extracted from it. However, such approaches are limited to the knowledge encoded in each sentence or paragraph, ignoring the knowledge that can be inferred from entire texts. 

A major obstacle for KBP progress has been the availability of datasets to train end-to-end KBP systems, mostly because it is too expensive to exhaustively annotate every possible triple that can be extracted from a large corpus \cite{chaganty2017importance}. 
NIST TAC KBP evaluation \cite{getman_laying_2018} covers text from Wikipedia, news, and social web, and datasets like KnowledgeNet \cite{mesquita2019knowledgenet} and WIKI \cite{trisedya-etal-2019-neural}, predominantly focus on general-purpose or encyclopedic knowledge bases like Wikidata. While useful in certain scenarios such datasets often fail to address the specific needs of domain-specific contexts where specialized vocabularies are required. 

\begin{figure}[t!]
    \centering
    \includegraphics[width=1\linewidth]{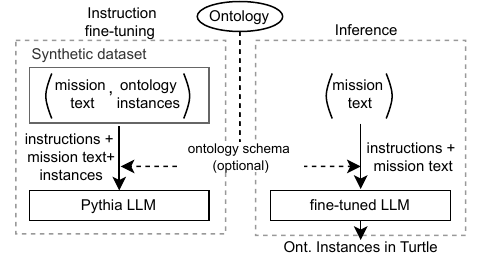}
    \caption{fine-tuning autoregressive LLM for KBP}
    \label{fig:approach}
\end{figure}

In this paper, we posit that it is possible to fine-tune a Large Language Model (LLM) for end-to-end KBP. The LLM analyzes the text, extracts the instances according to a predefined ontology, and formalizes them in a RDF serialization, such as Turtle (see figure \ref{fig:approach}). While LLM such as GPT4.0 are able to extract ontology instances from text and generate RDF serialization, our focus is on smaller models which, through supervised fine-tuning, can become specialists in KBP. Knowledge base curators can benefit from smaller LLM which are faster and consume less computational resources at inference time, implying also reduced costs. Our approach goes beyond state-of-the-art end-to-end models, supporting large input texts rather than sentences or paragraphs, and generating Turtle serialization of the extracted information. 

Our case study focuses on populating a space mission ontology \cite{Berquand2022Mission2KG}, tapping into a document corpus where space missions are described. We generate a KBP dataset for space missions using a state-of-the-art LLM and existing domain resources. Thus, we also contribute with a synthetic dataset for training and evaluation of end-to-end KBP in the space mission domain.\footnote{Code available at: \url{https://anonymous.4open.science/r/LLM4KBP-78FD}} In this scenario, we fine-tune the Pythia suite of language models \cite{biderman2023pythia} 
to investigate the relation between LLM size and the KBP task, demonstrating that it is possible to fit specialist models for KBP. The best performing model has 2.8B parameters, showing that in the scope of this KPB task, it is not necessary to train larger models. A 1B model is competitive with larger models of 6.9B and 12B models.

\section{Related work}
\textbf{Knowledge base population.}
Early KBP systems \cite{angeli2015leveraging, chaganty2017importance, mausam2016open} involve several subtasks such as named entity recognition, Coreference resolution, entity linking, and relation extraction \cite{orlando-etal-2024-relik} where errors can arise and propagate through the pipeline. 
End-to-end KBP systems \cite{mesquita2019knowledgenet} aim at populating knowledge bases in one step. End-to-end KBP relies on sequence to sequence approaches \cite{sutskever2014seq2seq} where an input sentence is translated into triples. \citet{stanovsky2018supervised} and \citet{miwa-bansal-2016-end} train biLSTM decoders to label the input sequence and obtain the triples. Other authors use the encoder-decoder framework that includes an autoregressive \cite{trisedya-etal-2019-neural,liu2018seq2rdf,huguet-cabot-navigli-2021-rebel-relation} or a non-autoregressive \cite{sui-etal-2021-set} decoder to generate the triples. Unlike current end-to-end models, we leverage a large pre-trained language model based on an autoregressive, decoder-only transformer \cite{radford2018improving}, which is further fine-tuned to generate all possible triples that conform to a predefined ontology from an input text, in a given RDF serialization. 

\textbf{Datasets.} 
The corpus in the NIS TAC KBP track \cite{getman_laying_2018} is annotated with a few general entity types and includes text from Wikipedia, news, blogs, and forums \cite{ji_knowledge_2011}. The TAC KBP 2020 RUFES track \cite{jiRUFES2020} covers news data and 200 entity type based on YAGO and Wikidata. KnowledgeNet \cite{mesquita2019knowledgenet} and WIKI \cite{trisedya-etal-2019-neural} are datasets intended to evaluate end-to-end KBP. Knowledgenet is manually annotated using 4 entity types, while WIKI is annotated using distance supervision covering 158 predicates. Both datasets are sentence-level, extracted from Wikipedia, and use Wikidata as KB. 

Datasets intended mainly for relation extraction, such as T-REx \cite{elsahar-etal-2018-rex}, fewRel \cite{han-etal-2018-fewrel} and TACRED \cite{alt-etal-2020-tacred} could be used to train KBP systems, although the fact that they are not exhaustively annotated makes them inadequate for end-to-end KBP evaluation. In contrast to existing KBP datasets focused on news, blogs, forums, and Wikipedia, our dataset is specific to the technological domain of space missions. Additionally, rather than providing individual sentences or paragraphs, our dataset includes comprehensive textual descriptions along with the complete RDF graph serialized in Turtle format according to a domain ontology.

\section{KBP Task formulation}
Let $O=(C,P,I)$ be an ontology\footnote{We constraint the ontology definition to a subset of RDF Schema \url{https://www.w3.org/TR/rdf-schema/}} where $C$ is the set of classes, $P$ is the set of properties, and $I$ the set of individuals. An individual $i \in I$ is any resource that belongs to a class in $C$. A triple $(s,p,o)$ states the relation $p \in P$ between a subject $s \in I$ and object $o \in I$. The type of subjects and objects for a given property can be restricted to a class or set of classes in $C$ through Domain and Range constraints. 

Let $d$ be a text document of the form $(w_1,w_2,...,w_n)$, where $w_j$ is its $j^{th}$ token. The KBP task identifies all triples $(s,p,o)$ in $d$ that satisfy the ontology $O$. That is, $p$ must be a valid property in the ontology ($p \in P$), $s$ belongs to the domain of $p$ $(Type(s) \in Domain(p))$, $o$ belongs to the range of $p$ $(Type(o) \in Range(p))$, and $s$ and $o$ are instances of classes in $C$ $(Type(s) \in C$ and $Type(o) \in C)$.

\section{Instruction tuning}
We fine-tune a pretrained autoregressive decoder-only transformer on the KBP task. These models follow the transformer architecture \cite{Vaswani2017AttentionIsALL} but keep only a decoder. The input is a sequence of tokens $x=(x_1,x_2,...,x_n)$. The model is trained, using cross-entropy loss, to learn the probability distribution over sequences by factorizing it in an autoregressive manner. 
\begin{align*}
 p(x)=p(x_1,x_2,...,x_n)=\prod_{t=1}^{n}p(x_t|x_{<t};\theta)
\end{align*}
In instruction tuning, the pretrained model learns to follow human instructions. To do so, the model is fine-tuned on instruction-response pairs using the same learning objective that in the pretraining stage. Let $X$ be the set of natural language instructions, $Y$ the set of responses, and $p_\theta(y|x)$ the language model parameterized by $\theta$, that learns the probability distribution over responses $y$ given an instruction $x$. 

For the task at hand, $x$ includes the instructions to generate the ontology individuals from an input text. The instruction also contains the input text, and optionally the ontology schema. The response $y$ is the expected Turtle serialization of the ontology individuals and their properties. 

\section{Resources}
We describe the resources used to generate a synthetic KBP dataset for the space mission domain. We also include resources useful for evaluating the generated dataset and the end-to-end KBP model. 

The \textbf{space mission ontology} \cite{Berquand2022Mission2KG} models space missions, their instruments, orbit, and stakeholders, as well as the countries involved in each mission (see table \ref{tab:mission-ont}). However, beyond the graphical representation, the ontology is not available in RDF/OWL format. Therefore, we reproduce and formalize the ontology using Protegé\footnote{\url{https://protege.stanford.edu}} to obtain a serialization in Turtle format.   

\begin{table}[]
\caption{Mission ontology: Classes and properties}
\label{tab:mission-ont}
\resizebox{\columnwidth}{!}{%
\begin{tabular}{@{}llll@{}}
\toprule
Class & Property & Type & Range/Literal \\
\midrule
\multirow{7}{*}{Mission} & missionName & Data property & xsd:string \\
 & missionStatus & Data property & xsd:string \\
 & launchDate & Data property & xsd:dateTime \\
 & endOfLife & Data property & xsd:dateTime \\
 & objectives & Data property & xsd:string \\
 & hasInstrument & Obj. property & Instrument \\
 & hasOrbit & Obj. property & Orbit \\
 \cmidrule{2-4}
\multirow{3}{*}{Instrument} & instrumentName & Data property & xsd:string \\
 & instrumentType & Data property & xsd:string \\
 & measurementsApp & Data property & xsd:string \\
 \cmidrule{2-4}
\multirow{3}{*}{Orbit} & orbitAltitude & Data property & xsd:string \\
 & orbitInclination & Data property & xsd:string \\
 & orbitType & Data property & xsd:string \\
 \cmidrule{2-4}
\multirow{4}{*}{Stakeholder} & stakeholderName & Data property & xsd:string \\
 & isBasedIn & Obj. property & Country \\
 & managesMission & Obj. property & Mission \\
 & ownsInstrument & Obj. property & Instrument \\
 \cmidrule{2-4}
Country & countryName & Data property & xsd:string \\
\bottomrule
\end{tabular}%
}
\end{table}

Maintained by the European Space Agency (ESA), \textbf{EOPortal} \footnote{\url{https://www.eoportal.org}}  contains more than 1,200 articles that fully describe space missions.  Descriptions usually contain an overview of the mission, the space segment, instruments and ground segments, as well as information about the launch and status of the mission. We use EOPortal mission description as input text to extract triples following the space mission ontology.

The database of the Commission on Earth Observation Satellites (\textbf{CEOS database} \footnote{\url{https://database.eohandbook.com/}}) provides information of space missions based on an annual survey of its member agencies. 
The CEOS database can be downloaded from the website in Excel format. In total, structured information on 725 missions and 1,055 instruments is downloaded. 172 missions in the CEOS database are described in EOPortal. We use the latter to evaluate both the generated training dataset and the KBP model output.

\section{Training data generation}
To fine-tune a LLM on the KBP task, we need a dataset with tuples containing the instructions to extract individuals from the space mission ontology, the input text describing the mission, and the expected output, which consists of the extracted individuals in Turtle format. The instructions are hand-crafted\footnote{the prompt is available at: \url{https://anonymous.4open.science/r/LLM4KBP-78FD}} and include a textual description of the space mission ontology. The input text is obtained from EOPortal.  

To generate the ontology individuals in Turtle format we use Llama3-8B-Instruct. This is a state-of-the-art LLM of 8 billion parameters\footnote{https://ai.meta.com/blog/meta-llama-3/} released by Meta in April 2024. While Llama3-8B has a large maximum sequence limit of 8,192 tokens, there are some mission descriptions in EOportal that largely exceed that limit
. Therefore, we summarize the missions descriptions using Llama3-8B. We truncate the mission description to fit within the maximum sequence limit and instruct the model to provide a summary in no more than four paragraphs, focusing on the main topics of the ontology, including mission information such as launch and end-of-life dates, objectives, orbit, instruments, and stakeholders. 

Then, we instruct Llama3-8B to extract the ontology individuals from the mission summary and formalize them in Turtle. The instructions are provided in a dialogue format: 

\begin{itemize}[noitemsep, leftmargin=*]
    \item \textit{user} asks the model to acknowledge that it understands the mission ontology that follows. The ontology is described in plain text.
    \item \textit{system} responds that it understands the ontology, and provides a brief summary.
    \item \textit{user} asks to extract instances of the ontology from the summary of a mission description. An example of the task is provided to the model.
    \item \textit{system} responds with the instances of the ontology that it extracts from the input text.
\end{itemize}





We use temperature$=$0.6 and top\_p$=$0.9 to generate the extracted instances in Turtle format. These parameter values were determined through previous experiments. However, Llama3-8B does not generate syntactically valid Turtles for the majority of input missions. Moreover, in some cases the triples do not respect the range and domain constraints of the properties. Therefore, we follow an iterative approach. First, we apply different heuristics at the lexico-syntactical and semantic level to attempt to fix the non-valid Turtle and make it complaint with the ontology schema. In the following generation iteration only missions with a non-valid Turtle are processed and the generated code is again syntactically and semantically evaluated and the heuristics applied. The process is repeated for a certain number of iterations. The heuristics we applied are as follows:

\begin{itemize}[noitemsep, leftmargin=*]  
    \item Lexical-syntactic corrections:
    \begin{itemize}[noitemsep, leftmargin=*] 
        \item Namespaces (NS): Replace all namespaces in the generated Turtle with the correct ones.
        \item Date format: Add default time to dates without time, elimination of triplets that include \textit{unknown} or empty dates in the object.
        \item Entities with a question mark prefix: The preceding question mark is removed from the individuals.
    \end{itemize}
    \item Semantic corrections:
    \begin{itemize}[noitemsep, leftmargin=*] 
        \item Domain constraints: All triples that do not comply with the domain restrictions expressed in the ontology are deleted.
        \item Range constraints: All triplets that do not comply with the range restrictions expressed in the ontology are deleted.
        \item Relations: Isolated instruments and orbits are automatically related to the mission entity.
        \item Replace subclass relations with rdf:type.
    \end{itemize}
\end{itemize}

The results of Turtle code generation for the instances extracted from the mission summaries are presented in Table \ref{table-gen}. From 1060 missions descriptions, we obtain syntactically valid Turtle respecting domain and range constraints for 1025 in 9 generation rounds. The heuristic with the greatest impact on turning the generated code into valid code is the overall replacement of the namespaces (NS). The final dataset consists of pairs of mission descriptions and individuals of the mission ontology in Turtle format.

\begin{table}[]
\caption{Statistics of the generation of training data. For each generation iteration (Iter.) we show the number of missions processed, the number of valid Turtles generated (Gen.) and after applying the different fixes.}
\label{table-gen}
\resizebox{\columnwidth}{!}{ 
\begin{tabular}{ccccccc}
\toprule
\multirow{2}{*}{Iter.} & \multirow{2}{*}{Missions} & \multicolumn{4}{c}{Valid Turtle}  & \multirow{2}{*}{\begin{tabular}[c]{@{}c@{}}Non-Valid \\ Turtle\end{tabular}} \\
\cmidrule(lr){3-6}
 &  & Gen. & NS & Lex-Sint. & Sem. &  \\
\midrule
1 & 1060 & 243 & 534 & 552 & 544 & 516 \\
2 & 516 & 97 & 208 & 221 & 220 & 296 \\
3 & 296 & 49 & 88 & 94 & 91 & 205 \\
4 & 205 & 26 & 42 & 46 & 46 & 159 \\
5 & 159 & 17 & 43 & 50 & 50 & 109 \\
6 & 109 & 12 & 23 & 25 & 25 & 84 \\
7 & 84 & 14 & 27 & 22 & 27 & 57 \\
8 & 57 & 7 & 11 & 13 & 13 & 44 \\
9 & 44 & 3 & 8 & 9 & 9 & 35 \\
\bottomrule
\end{tabular}
}
\end{table}

\section{Experiments}

As base models, we select the Pythia suite \cite{biderman2023pythia}. Pythia includes 8 LLM of variable size ranging from 14 million to 12 billion parameters, all trained on public data seen in exactly the same order. The models are trained on The Pile corpus, which covers different domains such as books, GitHub repositories, web pages, chat logs, and medical, physics, mathematics, computer science, and philosophy articles. We select models pre-trained on the deduplicated version of The Pile whenever available. The Pythia suite allows comparing the results of the fine-tuning process on the KBP task across different model sizes.  

To train the Pythia models on the KBP task for the space mission ontology we carry out additional supervised fine-tuning. The learning objective of the fine-tuning is the language modeling task, i.e., next token prediction, although in this case the text consists of pairs of instructions and responses. The instruction includes the ontology, and the mission description. The response is  the extracted individuals in Turtle format. From the dataset generated in the previous section we get a validation set that includes data from the 172 missions (17\% of the total) that have structured data in the CEOS database. Data from the remaining 853 missions (83\% of the total) are used for training.

We fine-tune the Pythia models in two scenarios: one with the ontology included in the instruction, and one without. All models are trained for 7 steps using an effective batch size of 8, learning rate  2e-5 and AdamW optimizer. We also apply early stopping when the loss function did not improve for two consecutive steps. To reduce the number of training parameters for models starting with 2.8B or more we use LoRA\cite{hu2021loralowrankadaptationlarge}. We setup LoRA with rank 16 and scaling factor($\alpha$) 32, and target the key value matrices of the attention blocks. 

\begin{figure*}[ht]
    \centering
    \includegraphics[width=1\linewidth]{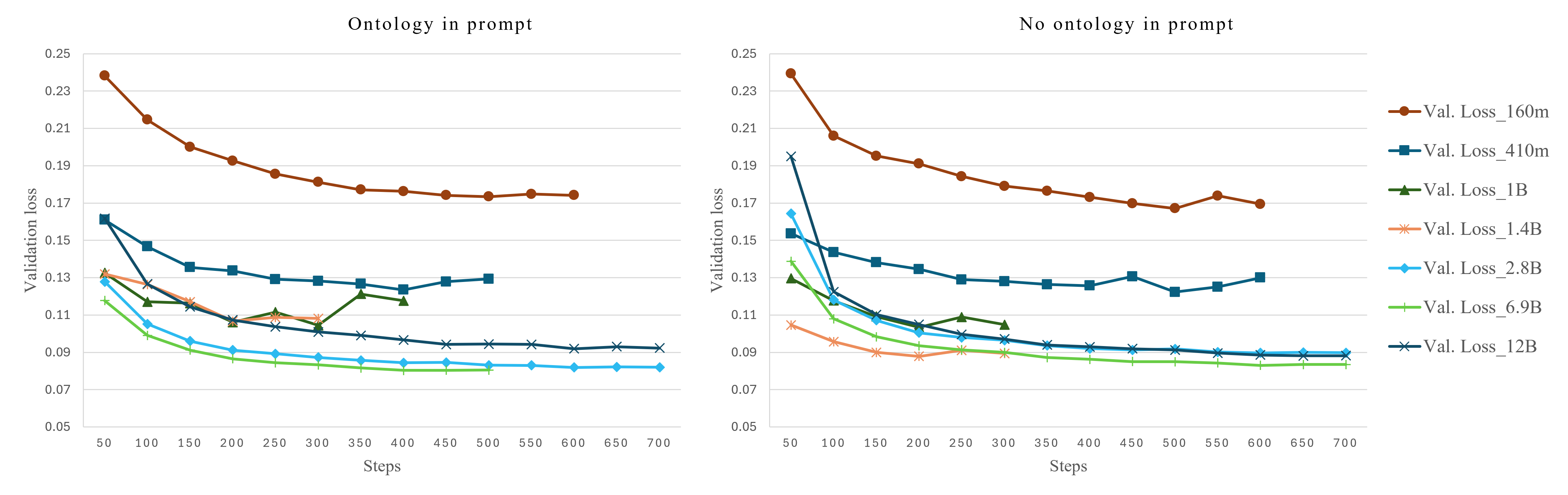}
    \caption{Validation loss at each step for the fine-tuned models. Models with less than 160m parameters are not shown since their loss is considerable higher. However, the validation loss for such models follows a pattern similar to that of the 160M model, suggesting that they have learned the task.}
    \label{fig:loss}
\end{figure*}

Training results are shown in Table \ref{table-eval1}. In general, we observe that as the model's parameters increase, perplexity and validation loss improve in the fine-tuned model. The exception is the 12B model, which has a higher validation loss than the 6.9B models, and also shows worse perplexity when the ontology is included in the instruction. At this stage is hard to decide the impact of including the ontology in the instructions beyond that validation loss improves only for 14m, 2.8B and 6.9B models. 
In Figure \ref{fig:loss}, we see that the training of the 1B and 1.4B models converges faster than the rest of the models. In general, models with more than 1B parameters generally have significantly lower loss than smaller ones. The 6.9B model achieves the lowest loss in both scenarios (with or without ontology in the instruction).

\begin{table}[htb!]
\caption{Evaluation loss and perplexity of pythia models fine-tuned on the KBP task.}
\label{table-eval1}
\resizebox{\columnwidth}{!}{ 
\begin{tabular}{@{}ccccccccc@{}}
\toprule
\multirow{2}{*}{Model}  & \multirow{2}{*}{LoRA} & \multicolumn{2}{c}{Ont. in prompt} & \multicolumn{2}{c}{No ont. in prompt} \\ \cmidrule(lr){3-4} \cmidrule(lr){5-6} 
 &   &  Perplexity & Loss & Perplexity & Loss \\ \midrule
14m  & No & 1.35 & 0.297 & 1.35 & 0.303 \\
31m  & No & 1.30 & 0.261 & 1.30 & 0.259 \\
70m  & No & 1.26 & 0.231 & 1.25 & 0.225 \\
160m  & No & 1.19 & 0.173 & 1.18 & 0.167 \\
410m  & No & 1.13 & 0.123 & 1.13 & 0.122 \\
1B  & No  & 1.11 & 0.104 & 1.11 & 0.103 \\
1.4B  & No & 1.11 & 0.107 & 1.11 & 0.103 \\
2.8B  & Yes & 1.09 & 0.082 & 1.09 & 0.090 \\
6.9B  & Yes & 1.08 & 0.080 & 1.09 & 0.083 \\
12B  & Yes  & 1.10 & 0.092 & 1.09 & 0.088 \\ \bottomrule
\end{tabular}
}
\end{table}

\subsection{Evaluation}
\begin{figure}[htb]
    \centering
    \includegraphics[width=1\linewidth]{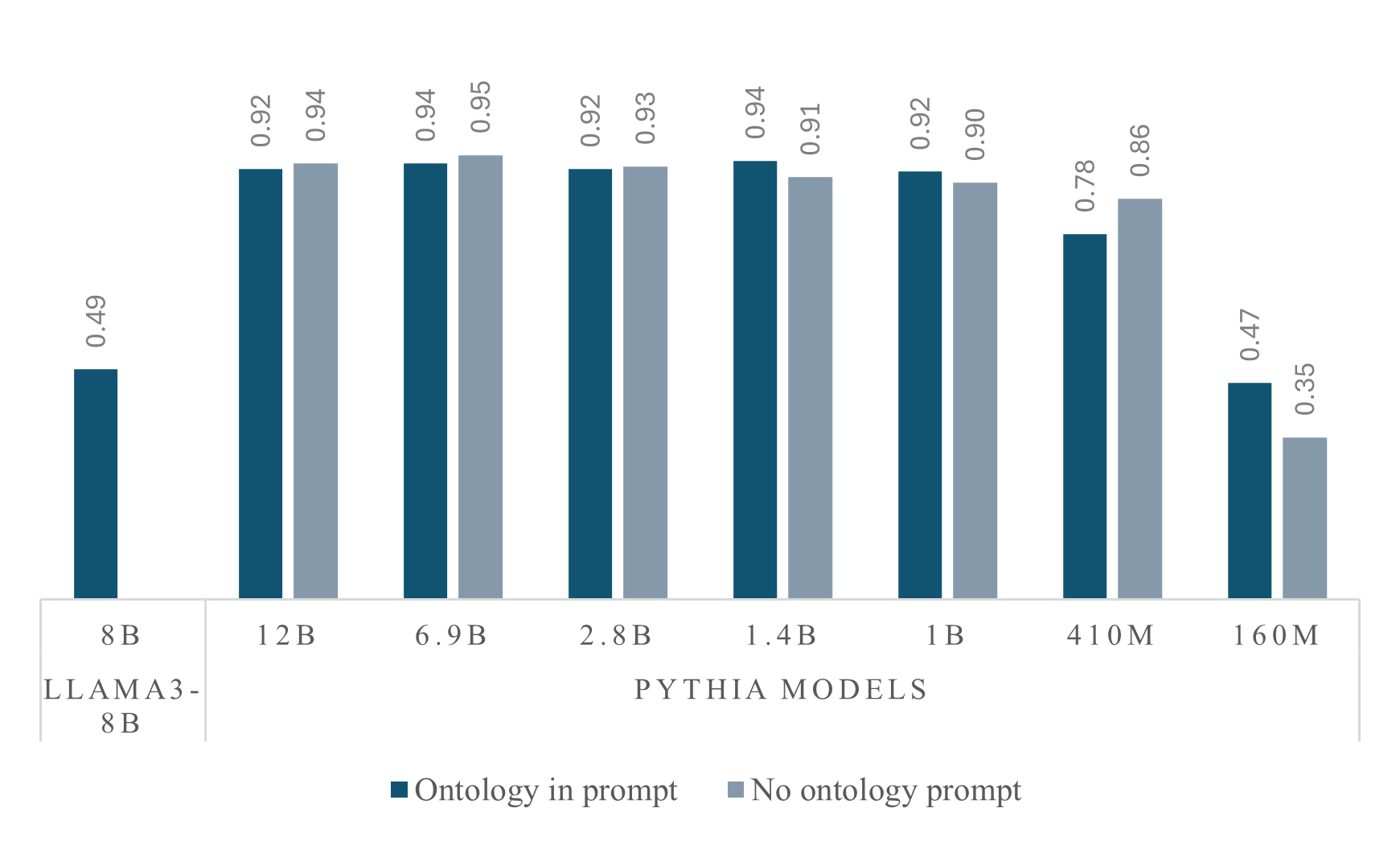}
    \caption{Syntactically valid Turtle generated from the validation set by the fine-tuned models. We show the result of llama3-8B (zero-shot) as a baseline.}
    \label{fig:ValidTurtle}
\end{figure}

We evaluate the fine-tuned models to validate that the generated Turtle format is syntactically correct and that the ontology individuals extracted from the text correspond to the input mission. Figure \ref{fig:ValidTurtle} shows the percentage of syntactically valid Turtle generated by the fine-tuned models tuned when they process the validation dataset. Small models with 14m and 70m parameters are unable to generate correct Turtle. Starting from 160m parameters, the models generate syntactically valid Turtle. If we compare the Pythia models tuned with llama3-8B in zero-shot, i.e. without fine-tuning on the task, we observe that the models with 410m parameters or more generate a higher percentage of valid Turtle. Starting with 1B parameters, the fine-tuned models generate 90\% or more of syntactically correct Turtle code. This is evidence that the fine-tuned models have learned the task and do it better than a state-of-the-art 8B model. Furthermore, including the ontology in the prompt slightly hurts models with 2.8B parameters or more, while it benefits models with 1.4B and 1B parameters.

We use structured information from the CEOS database to semantically evaluate the generation of the fine-tuned models in the validation set. We issue SPARQL queries on the valid Turtle generated by the models to identify the literal values or objects of the properties of the ontology individuals. The value of each property is compared with the value of that property in the corresponding attribute in CEOS. As evaluation metrics we use Rouge-L, which is based on the longest common subsequence (LCS) between the reference and generated text. Rouge-L is adequate to compare text of similar length. However when the compared text significantly differs in length, e.g. a summary and the source text, Rouge-L performs poorly. Thus, to complement the evaluation we use a LLM as a judge. We instruct LLama3-8B to generate a semantic similarity score between 0 and 1. 

\begin{figure*}[ht]
    \centering
    \includegraphics[width=1\linewidth]{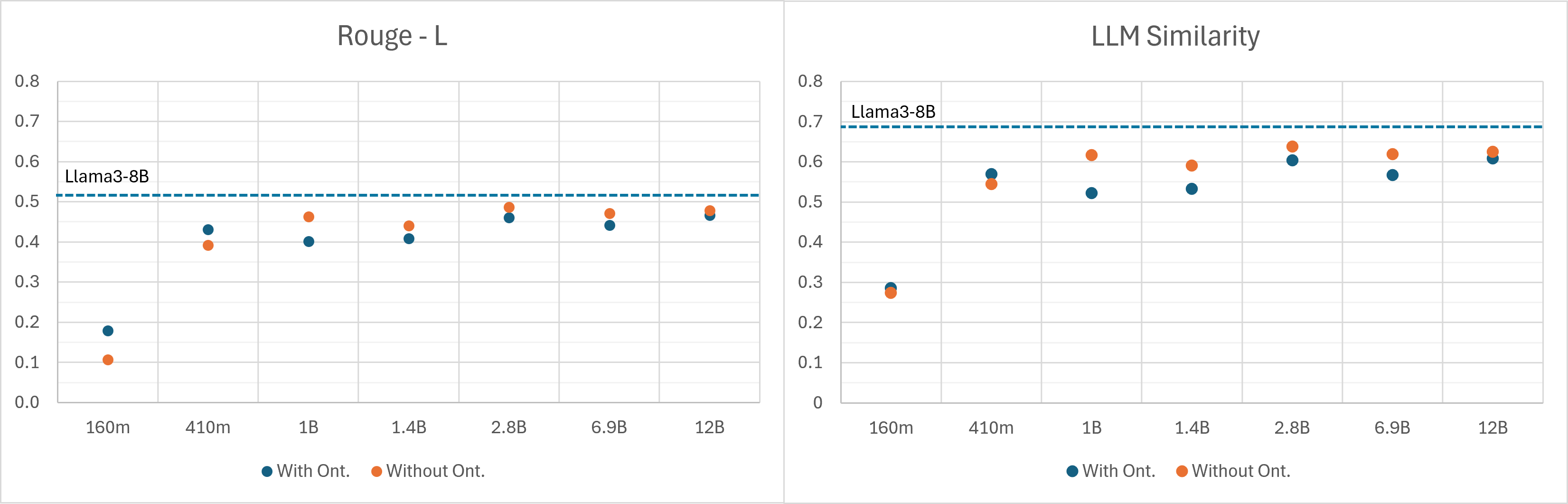}
    \caption{Evaluation of Pythia models trained including or not the ontology in the prompt according to the Rouge-L  and the LLM-based similarity.}
    \label{fig:llmeval}
\end{figure*}



Evaluation results of the two best models are shown in table \ref{tab:sem_eval} (complete results available in appendix \ref{sec:appendix}). The best pythia models, 2.8B and 12B in that order, are fine-tuned without including the ontology in the instructions. Llama3-8B zero-shot is included in the table as upper bound since the models are fine-tuned on synthetic data generated using this model. However, note that Llama3-8B only generates valid turtle for 49\% of the missions, while Pythia models with 1B o more parameters produce more than 90\% valid turtle. For properties such as \textit{Mission objective} or \textit{Applications of instrument measurements}, Rouge-L produces very heterogeneous and low scores compared to those of the LLM-based similarity metric. This is because the generation of the fine-tuned model is much more concise than the expected responses in the CEOS database. LLM-based similarity yields higher scores because the judge considers the semantics of texts and does not rely so much on the way they are written. A hard property to predict accurately is the end-of-life date (EOL date). We observed occasional mismatches between EOL dates in EOPortal mission descriptions and the CEOS database, as the former lists planned dates while the latter records actual dates.

\begin{table}[ht!]
\caption{Evaluation of property values according to expected values in the CEOS database. Best fine-tuned models do not include the ontology in the prompt.}
\label{tab:sem_eval}
\resizebox{\columnwidth}{!}{%
\begin{tabular}{@{}lcccccc@{}}
\toprule
\multicolumn{1}{c}{\multirow{2}{*}{Model}} & \multicolumn{2}{c}{Llama3} & \multicolumn{4}{c}{Pythia (fine-tuned)} \\
\cmidrule(lr){2-3} \cmidrule(lr){4-7} 
\multicolumn{1}{c}{} & \multicolumn{2}{c}{8B} & \multicolumn{2}{c}{2.8B} & \multicolumn{2}{c}{12B} \\ 
\midrule
Properties/metrics & Rouge & Sim. & Rouge & Sim. & Rouge & Sim. \\
\midrule
Mission Name & 0.924 & 0.949 & 0.845 & 0.894 & 0.864 & 0.902 \\
Mission Status & 0.515 & 0.685 & 0.603 & 0.691 & 0.563 & 0.657 \\
Launch Date & 0.668 & 0.738 & 0.661 & 0.698 & 0.670 & 0.704 \\
EOL Date & 0.226 & 0.325 & 0.241 & 0.311 & 0.301 & 0.342 \\
Objectives & 0.156 & 0.669 & 0.143 & 0.616 & 0.139 & 0.622 \\
Orbit Inclination & 0.607 & 0.795 & 0.568 & 0.674 & 0.513 & 0.628 \\
Orbit Altitude & 0.759 & 0.786 & 0.633 & 0.661 & 0.607 & 0.637 \\
Orbit Type & 0.742 & 0.677 & 0.508 & 0.530 & 0.496 & 0.520 \\
Instrument Name & 0.460 & 0.627 & 0.540 & 0.687 & 0.533 & 0.628 \\
Measurements app. & 0.160 & 0.696 & 0.148 & 0.676 & 0.160 & 0.678 \\
Stakeholder Name & 0.413 & 0.561 & 0.468 & 0.581 & 0.416 & 0.559 \\ \midrule
\multicolumn{1}{r}{Average} & 0.512 & 0.683 & 0.487 & 0.638 & 0.478 & 0.625 \\\bottomrule
\end{tabular}%
}
\end{table}

Figure \ref{fig:llmeval} shows the average evaluation metrics for the fine-tuned models over all the evaluated properties. Note that starting from 1B parameters, it is always better to exclude the ontology from the prompt. This is a benefit because a shorter prompt is processed faster and requires less GPU memory, which translates into cost savings. Fine-tuned models with 160m parameters or less do not learn to extract property values of the mission ontology. While the best models according to both metrics are the largest ones, 2.8B, 6.9B and 12B, the 1B model is already competitive and could be a valid option in low computing resource scenarios. 


\section{Conclusions}
In this paper we posit that it is possible to fine-tune autoregressive language models for end-to-end KBP. Our approach allows processing large input text all at once, and generates not only triples but ontology individuals serialized in Turtle format. We experiment with a domain ontology of space missions, which contrasts with the usual general-purpose knowledge bases used in the KBP literature. 
A key contribution of this paper is a synthetic dataset for end-to-end KBP in the space mission domain. 
We find that fine-tuned models starting from 410m parameters learn to extract ontology individuals from input text and formalize them in an RDF serialization. For models with 1B parameters or more, is better not include the ontology in the prompt, while models with lesser parameters are enhanced when the ontology is included. Overall the best model is a 2.8B model which highlights that for this KBP scenario is not necessary to use larger language models that requires more computational resources to run. In fact, the model with 1B parameters is already very competitive with the largest 6.9B and 12B versions.

\section*{Limitations}
Our approach to KBP requires models to process input text and generate RDF serializations that can be potentially very long. While pythia models have a max sequence length of 2048 which is enough for our use case, other larger input text and ontologies might be a problem. Nevertheless, the current trend is that larger context lengths are being supported, e.g., Llama3 and Llama3.1 models have a context length of 8K tokens and 128K tokens. Another limitation is that our approach does not cover entity canonicalization, i.e., mapping entity mentions to their proper entity ID in a KB, which is necessary to avoid adding duplicate individuals to the knowledge base. We will address canonicalization in future work.





\bibliography{anthology,custom}
\bibliographystyle{acl_natbib}

\appendix
\section{Fine-tuned models evaluated in space mission ontology}\label{sec:appendix}

\begin{landscape}
\begin{table}[]
\centering
\caption{Rouge-L of Pythia models trained with and without the ontology in the prompt.}
\label{tab:evalpythianoont}
\resizebox{0.7\linewidth}{!}{%
\begin{tabular}{@{}cccccccccccccccc@{}}
\toprule
\multirow{3}{*}{Property} & \multirow{2}{*}{Llama 3} & \multicolumn{14}{c}{Pythia Models} \\ \cmidrule(l){3-16} 
 &  & \multicolumn{2}{c}{12B} & \multicolumn{2}{c}{6.9B} & \multicolumn{2}{c}{2.8B} & \multicolumn{2}{c}{1.4B} & \multicolumn{2}{c}{1B} & \multicolumn{2}{c}{410m} & \multicolumn{2}{c}{160m} \\ \cmidrule(lr){2-2} \cmidrule(l){3-16}
 & Ont. & Ont. & No Ont. & Ont. & No Ont. & Ont. & No Ont. & Ont. & No Ont. & Ont. & No Ont. & Ont. & No Ont. & Ont. & No Ont. \\ \cmidrule(r){1-1} \cmidrule(lr){2-2} \cmidrule(l){3-16}
Mission Name & 0.924 & 0.843 & 0.864 & 0.781 & 0.820 & 0.804 & 0.845 & 0.695 & 0.764 & 0.636 & 0.810 & 0.732 & 0.694 & 0.330 & 0.135 \\
Mission Status & 0.515 & 0.552 & 0.563 & 0.472 & 0.562 & 0.516 & 0.603 & 0.464 & 0.484 & 0.389 & 0.557 & 0.455 & 0.361 & 0.061 & 0.020 \\
Launch Date & 0.668 & 0.639 & 0.670 & 0.600 & 0.652 & 0.627 & 0.661 & 0.534 & 0.597 & 0.502 & 0.571 & 0.566 & 0.377 & 0.029 & 0.000 \\
EOL Date & 0.226 & 0.260 & 0.301 & 0.206 & 0.271 & 0.207 & 0.241 & 0.251 & 0.253 & 0.214 & 0.258 & 0.165 & 0.111 & 0.000 & 0.000 \\
Objectives & 0.156 & 0.136 & 0.139 & 0.122 & 0.140 & 0.136 & 0.143 & 0.127 & 0.138 & 0.115 & 0.136 & 0.134 & 0.127 & 0.059 & 0.028 \\
Orbit Inclination & 0.607 & 0.529 & 0.513 & 0.481 & 0.523 & 0.544 & 0.568 & 0.444 & 0.529 & 0.489 & 0.547 & 0.485 & 0.522 & 0.259 & 0.065 \\
Orbit Altitude & 0.759 & 0.619 & 0.607 & 0.535 & 0.581 & 0.633 & 0.633 & 0.540 & 0.592 & 0.538 & 0.606 & 0.595 & 0.595 & 0.311 & 0.115 \\
Orbit Type & 0.742 & 0.506 & 0.496 & 0.484 & 0.504 & 0.557 & 0.508 & 0.477 & 0.503 & 0.467 & 0.549 & 0.519 & 0.526 & 0.142 & 0.110 \\
Instrument Name & 0.460 & 0.513 & 0.533 & 0.526 & 0.531 & 0.493 & 0.540 & 0.478 & 0.432 & 0.433 & 0.507 & 0.503 & 0.381 & 0.407 & 0.276 \\
Instrument Application & 0.160 & 0.154 & 0.160 & 0.178 & 0.161 & 0.149 & 0.148 & 0.152 & 0.153 & 0.138 & 0.157 & 0.155 & 0.171 & 0.122 & 0.187 \\
Stakeholder Name & 0.413 & 0.385 & 0.416 & 0.470 & 0.441 & 0.400 & 0.468 & 0.339 & 0.400 & 0.500 & 0.390 & 0.437 & 0.441 & 0.256 & 0.241 \\ \midrule
Average & 0.512 & 0.467 & 0.478 & 0.442 & 0.471 & 0.461 & 0.487 & 0.409 & 0.440 & 0.402 & 0.463 & 0.432 & 0.391 & 0.180 & 0.107 \\ \bottomrule
\end{tabular}%
}
\end{table}

\begin{table}[]
\centering
\caption{Evaluation using LLM-similarity of Pythia models with and without including the ontology in the prompt.}
\label{tab:evalpythiaLLMsim}
\resizebox{0.7\linewidth}{!}{%
\begin{tabular}{@{}cccccccccccccccc@{}}
\toprule
\multirow{3}{*}{Property} & \multirow{2}{*}{Llama 3} & \multicolumn{14}{c}{Pythia Models} \\ \cmidrule(l){3-16} 
 &  & \multicolumn{2}{c}{12B} & \multicolumn{2}{c}{6.9B} & \multicolumn{2}{c}{2.8B} & \multicolumn{2}{c}{1.4B} & \multicolumn{2}{c}{1B} & \multicolumn{2}{c}{410m} & \multicolumn{2}{c}{160m} \\ \cmidrule(lr){2-2} \cmidrule(l){3-16}
 & Ont. & Ont. & No Ont. & Ont. & No Ont. & Ont. & No Ont. & Ont. & No Ont. & Ont. & No Ont. & Ont. & No Ont. & Ont. & No Ont. \\ \cmidrule(r){1-1} \cmidrule(lr){2-2} \cmidrule(l){3-16}
Mission Name & 0.949 & 0.851 & 0.902 & 0.790 & 0.876 & 0.823 & 0.894 & 0.718 & 0.826 & 0.655 & 0.876 & 0.766 & 0.769 & 0.372 & 0.364 \\
Mission Status & 0.685 & 0.678 & 0.657 & 0.542 & 0.671 & 0.639 & 0.691 & 0.554 & 0.606 & 0.458 & 0.653 & 0.559 & 0.512 & 0.071 & 0.037 \\
Launch Date & 0.738 & 0.666 & 0.704 & 0.640 & 0.661 & 0.661 & 0.698 & 0.556 & 0.602 & 0.532 & 0.635 & 0.562 & 0.345 & 0.029 & 0.000 \\
EOL Date & 0.325 & 0.299 & 0.342 & 0.261 & 0.349 & 0.284 & 0.311 & 0.273 & 0.293 & 0.250 & 0.349 & 0.218 & 0.162 & 0.000 & 0.000 \\
Objectives & 0.669 & 0.573 & 0.622 & 0.507 & 0.624 & 0.538 & 0.616 & 0.480 & 0.621 & 0.444 & 0.623 & 0.527 & 0.643 & 0.260 & 0.697 \\
Orbit Inclination & 0.795 & 0.643 & 0.628 & 0.572 & 0.636 & 0.664 & 0.674 & 0.536 & 0.630 & 0.575 & 0.653 & 0.605 & 0.614 & 0.351 & 0.088 \\
Orbit Altitude & 0.786 & 0.630 & 0.637 & 0.575 & 0.609 & 0.663 & 0.661 & 0.563 & 0.617 & 0.561 & 0.634 & 0.618 & 0.623 & 0.343 & 0.138 \\
Orbit Type & 0.677 & 0.514 & 0.520 & 0.474 & 0.525 & 0.550 & 0.530 & 0.453 & 0.535 & 0.450 & 0.558 & 0.511 & 0.550 & 0.153 & 0.195 \\
Instrument Name & 0.627 & 0.647 & 0.628 & 0.646 & 0.644 & 0.629 & 0.687 & 0.608 & 0.593 & 0.587 & 0.641 & 0.644 & 0.580 & 0.525 & 0.423 \\
Instrument Application & 0.696 & 0.655 & 0.678 & 0.666 & 0.645 & 0.657 & 0.676 & 0.620 & 0.657 & 0.620 & 0.645 & 0.676 & 0.624 & 0.611 & 0.617 \\
Stakeholder Name & 0.561 & 0.538 & 0.559 & 0.572 & 0.569 & 0.536 & 0.581 & 0.500 & 0.526 & 0.618 & 0.523 & 0.585 & 0.564 & 0.431 & 0.451 \\  \midrule
Average & 0.683 & 0.609 & 0.625 & 0.568 & 0.619 & 0.604 & 0.638 & 0.533 & 0.591 & 0.523 & 0.617 & 0.570 & 0.544 & 0.286 & 0.274 \\ \bottomrule
\end{tabular}%
}
\end{table}

\end{landscape}

\end{document}